\documentclass{optica-article}

\journal{opticajournal} 

\articletype{Research Article}

\usepackage{lineno}
\usepackage{graphicx}%
\usepackage{multirow}%
\usepackage{mathrsfs}%
\usepackage[title]{appendix}%
\usepackage{xcolor}%
\usepackage{textcomp}%
\usepackage{manyfoot}%
\usepackage{booktabs}%
\usepackage{algorithm}%
\usepackage{algorithmicx}%
\usepackage{algpseudocode}%
\usepackage{listings}%
\usepackage{array}%
\usepackage{multirow}%
\usepackage{booktabs}%
\usepackage{siunitx}

\begin{document}
	
\title{Long-Distance Field Demonstration of Imaging-Free Drone Identification in Intracity Environments}

\author{Junran Guo,\authormark{1,\dag} Tonglin Mu,\authormark{1,\dag} Keyuan Li,\authormark{1} Jianing Li,\authormark{1} Ziyang Luo,\authormark{1} Ye Chen,\authormark{1} Xiaodong Fan,\authormark{1} Jinquan Huang,\authormark{1} Minjie Liu,\authormark{1} Jinbei Zhang,\authormark{1} Ruoyang Qi,\authormark{2,*} Naiting Gu, \authormark{3,**} and Shihai Sun\authormark{1,***}}

\address{\authormark{1}School of Electronics and Communication Engineering, Sun Yat-sen University, Shenzhen, 518107, Guangdong, P.R.China\\
	\authormark{2} China Academy of Space Technology Beijing, 100094, Beijing, P.R.China\\
	\authormark{3} College of Advanced Interdisciplinary Studies, National University of Defense Technology, Changsha 410073, China\\
	\authormark{\dag}The authors contributed equally to this work.\\
	\authormark{*}qry15@tsinghua.org.cn\\
	\authormark{**}gnt7328@163.com\\
	\authormark{***}sunshh8@mail.sysu.edu.cn} 


\begin{abstract*} 
	Detecting small objects, such as drones, over long distances presents a significant challenge with broad implications for security, surveillance, environmental monitoring, and autonomous systems. Traditional imaging-based methods rely on high-resolution image acquisition, but are often constrained by range, power consumption, and cost. In contrast, data-driven single-photon-single-pixel light detection and ranging (\text{D\textsuperscript{2}SP\textsuperscript{2}-LiDAR}) provides an imaging-free alternative, directly enabling target identification while reducing system complexity and cost. However, its detection range has been limited to a few hundred meters. Here, we introduce a novel integration of residual neural networks (ResNet) with \text{D\textsuperscript{2}SP\textsuperscript{2}-LiDAR}, incorporating a refined observation model to extend the detection range to 5~\si{\kilo\meter} in an intracity environment while enabling high-accuracy identification of drone poses and types. Experimental results demonstrate that our approach not only outperforms conventional imaging-based recognition systems, but also achieves 94.93\% pose identification accuracy and 97.99\% type classification accuracy, even under weak signal conditions with long distances and low signal-to-noise ratios (SNRs). These findings highlight the potential of imaging-free methods for robust long-range detection of small targets in real-world scenarios.
	
\end{abstract*}

\section{Introduction}\label{sec1}

The proliferation of drones has transformed the applications in air defense \cite{rossiter2018drone}, search and rescue \cite{mishra2020drone}, and security \cite{dilshad2020applications}, enabling rapid response and enhancing situational awareness. However, their increasing application in large-scale and complex environments presents significant challenges, particularly in long-range detection and tracking. The ability to reliably identify drones over extended distances is critical for both civilian and defense applications, yet existing detection technologies remain constrained by limitations in range, power efficiency, and cost.

\begin{figure}[H]
	\centering
	\includegraphics[width=0.8\textwidth]{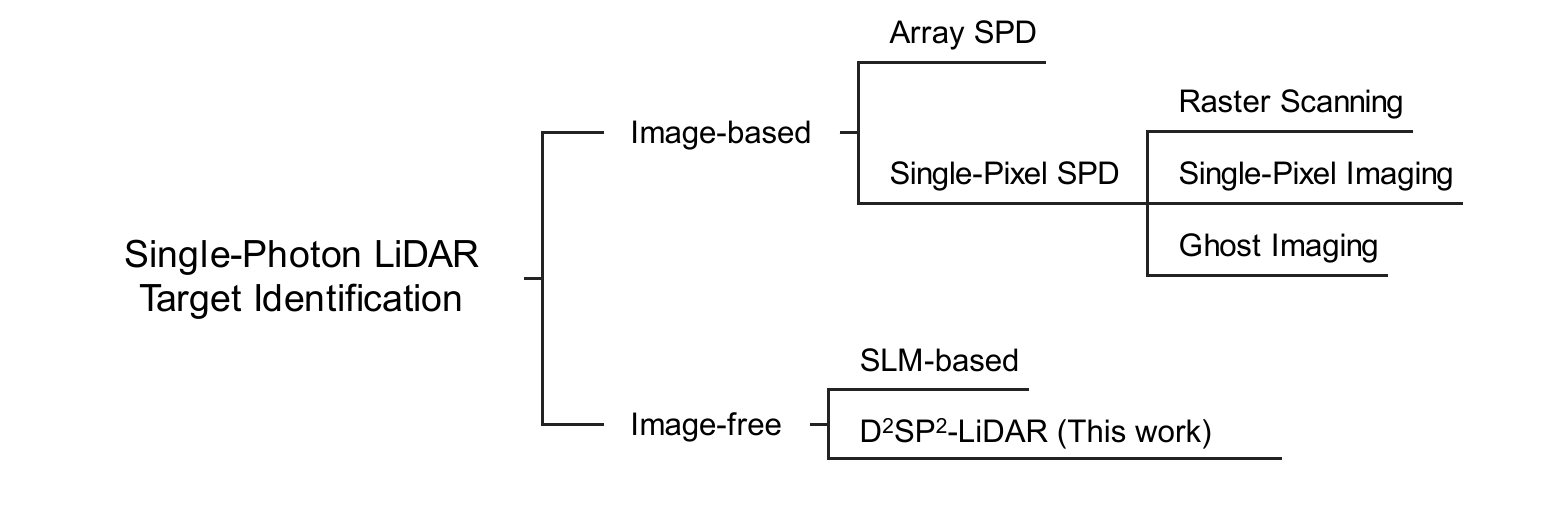}
	\caption{Conclusion of single-photon LiDAR target identification methods.}\label{fig1}
\end{figure}

Vision imaging \cite{lee2018drone,taha2019machine,haddad2020long,zhang2021autonomous}, such as RGB or intensity images, is limited in long-range scenarios. An extremely long focal length is required to prevent the decrease in transverse resolution caused by the diffraction limit. For radar systems \cite{coluccia2020detection}, such as phased array \cite{tang2019small, feng2019design, yang2021cylindrical}, frequency-modulated continuous-wave radars \cite{rahman2020classification, roldan2020dopplernet, hanif2022micro, sayed2024frequency} or pulse Doppler radars \cite{wang2021deep, gong2022detection, tian2024fully}, high power consumption and cost are main bottlenecks for long detection ranges. Single-photon light detection and ranging (LiDAR) technology has attracted significant attention recently due to its extended operating range and high accuracy in target identification \cite{pawlikowska2017single, li2020single,li2021single}. As shown in Fig. \ref{fig1}, existing single-photon LiDAR target identification methods can be mainly divided into image-based and imaging-free approaches. For image-based approaches, depth or reflectivity images are first reconstructed with array single-photon detectors (SPD) or single-pixel SPD, and then the targets are identified with the help of deep learning \cite{mora2021high, scholes2022dronesense}. Array-SPD systems \cite{tachella2019real, chan2019long, scholes2022dronesense, mora2021high, jiang2023long, mora2024human} face physical constraints imposed by the diffraction limit, and a high definition SPD array still presents significant technological challenges and high costs. Utilizing raster scanning~\cite{pawlikowska2017single, li2020single,li2021single}, computational single-pixel imaging (SPI) \cite{duarte2008single, edgar2019principles, huang2022scanning, song2023single}, and ghost imaging (GI) \cite{zou2023target,ruget2024translated, abbas2025target}, it is possible to reconstruct images with single-pixel SPD. It suffers from some limitations. The raster-scanning method is limited by the integration time per pixel, which restricts the detection of fast-moving dynamic targets. On the other hand, SPI or GI require extra components such as spatial light modulators (SLMs) or quantum entangled light sources, which introduce high system complexity and increased cost. At the same time, in the imaging-free methods, the targets are directly identified and detected from the measured data captured by the single-pixel SPD. Although imaging-free methods have been demonstrated recently based on SPI \cite{lohit2016direct, latorre2019online, zhang2020image, zhu2020photon, peng2023image, guo2024real, yu2024long} and GI \cite{hualong2021non, he2021handwritten, cao2021single}, they still face the same issues of increased cost and system complexity due to their reliance on SLMs for active modulation.

\begin{figure}[H]
	\centering
	\includegraphics[width=0.9\textwidth]{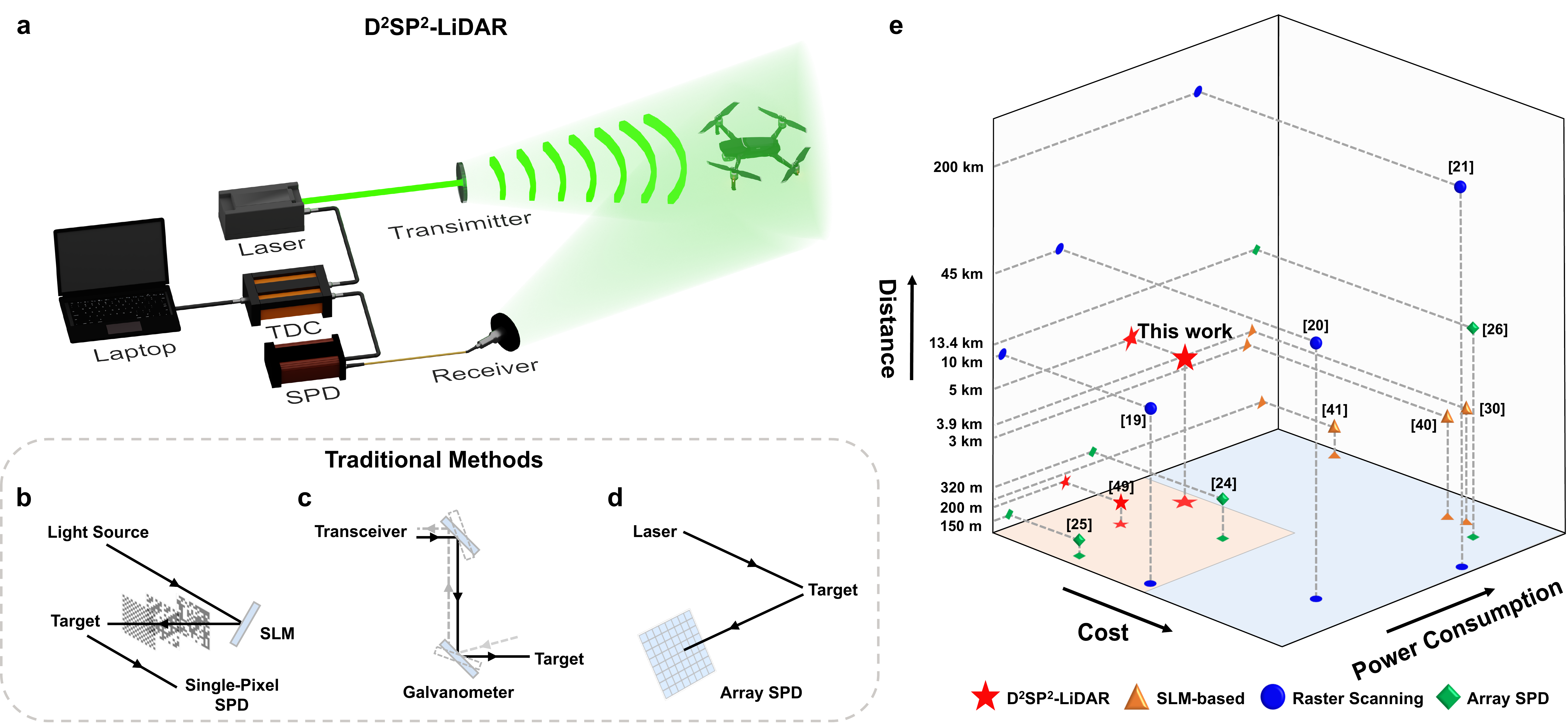}
	\caption{\textbf{a} Demonstration of \text{D\textsuperscript{2}SP\textsuperscript{2}-LiDAR}. \textbf{b} Target identification using single-pixel imaging. \textbf{c} Raster scanning imaging using galvanometer. The transceiver is composed of a pulsed laser and a single-pixel SPD. \textbf{d} Imaging using array SPD. \textbf{e} Comparison of distance, cost, and power consumption of different single-photon target detection methods. Our work shows low cost and low-power consumption while maintaining high accuracy in long-range scenarios.}\label{fig2}
\end{figure}

Therefore, implementing drone detection and identification with long-range capabilities, low cost, and low-power consumption remains a major challenge. Recently, a data-driven single-photon-single-pixel LiDAR (\text{D\textsuperscript{2}SP\textsuperscript{2}}-LiDAR)  has drawn much attention \cite{caramazza2018neural, turpin2020spatial, turpin20213d, liang2022non, hong2023image,lai2024single} due to its ability to capture effective information of the target without additional imaging components and modulators, such as SLMs (Fig. \ref{fig2} \textbf{b}), galvanometers (Fig. \ref{fig2} \textbf{c}), or array sensors (Fig. \ref{fig2} \textbf{d}). Research studies have shown that with \text{D\textsuperscript{2}SP\textsuperscript{2}}-LiDAR, one can reconstruct the depth image of the scene \cite{turpin2020spatial, turpin20213d, liang2022non, lai2024single}, identify and locate humans behind walls \cite{caramazza2018neural}, and discriminate different poses and types of drones at a distance of 200~\si{\meter} \cite{hong2023image}, revealing a promising low-cost target identification approach.

Here, to further improve the distance and accuracy of small target detection, we present an imaging-free and data-driven approach that integrates \text{D\textsuperscript{2}SP\textsuperscript{2}-LiDAR} (Fig. \ref{fig2} \textbf{a}) with residual neural networks (ResNet), achieving long-range drone detection and classification without conventional imaging components. Our method bypasses the limitations of extra components and directly extracts the features of the target from time-of-flight (ToF) data. And a field experiment in an intracity environment is demonstrated, which significantly extends the detection range from the previous limits of a few hundred meters to 5 \si{\kilo\meter}, while maintaining a high classification accuracy of 97.99\% for drone types and 94.93\% for drone pose identification. Both simulation and experimental results confirm the robustness of our method under weak signal conditions, highlighting its potential for real-world applications in long-range surveillance and autonomous systems.

\section{Methods}\label{sec2}
\subsection{Observation Model}\label{subsec21}
Consider the detection approach in Fig. \ref{fig2} \textbf{a}, the target is illuminated by a pulsed laser with temporal shape $s(t)$, and the echo photons are collected by a single-pixel SPD which is synchronized to the laser with a time digital converter (TDC). The echo signal of the target can be described by convolving $s(t)$ with the target response function $h(t)=\sum_ia_i\delta(t-2d_i/c)$. Here, $a_i$ is the falloff effect coefficient, $\delta(t-2d_i/c)$ represents the round-trip time shift of the echo signal, $c$ is the speed of light, and $i$ denotes the index of the target reflect surface with $d_i$ being the distance between the surface of object and the \text{D\textsuperscript{2}SP\textsuperscript{2}-LiDAR} system. Combining with $J(t)$, the time jitter of the system, we can obtain the echo signal $e(t)$ incidents on the SPD
\begin{equation}
	e(t)=J(t)*s(t)*h(t)+B_b.
	\label{eq1}
\end{equation}
Here $*$ represents the convolution operation. $B_b$ denotes background noise photons originating from sunlight and urban lighting, which follows a uniform distribution in time \cite{rapp2017few}. We first consider the period in a single pulse duration. Therefore, the mean photon number of the echo signal detected in the $k$-th bin can be expressed as
\begin{equation}
	N_k=\int_{k\Delta t}^{(k+1)\Delta t}(\eta e(t)+B_d)dt=\int_{k\Delta t}^{(k+1)\Delta t}\eta e(t)dt+B,\label{eq2}
\end{equation}
where $\eta$ is the detection efficiency of the SPD, $B_d$ is the dark count of the SPD, $\Delta t$ is the width of time bin, $B$ represents noise containing $B_d$ and $B_b$. Ideally, considering the detection process of SPD, the detection probability of each time bin of the temporal histogram follows Poisson distribution \cite{gatt2009geiger} in each time bin, the probability that at least one photon is detected in the $k$-th bin is given by
\begin{equation}
	p_0(k)=1-exp(-N_k).\label{eq3}
\end{equation}
However, because the system isn’t working in the low-flux regime \cite{rapp2017few}, where the average count rate is limited to at most 1\%$\sim$5\% of the illumination periods, the dead time $t_d$ of the SPD needs to be considered. Then the number of bins corresponding to the dead time can be calculated as $\text{bin}_d=[t_d/\Delta t]$, where $[\cdot]$ is a rounding operation. Then the detection probability should be rewritten in a recursive form \cite{li2017influence}
\begin{equation}
	P(k) =
	\begin{cases}
		p_0(k)\left[1 - \sum_{i=k-\text{bin}_d}^{k-1} P(i)\right], & k \geq \text{bin}_d \\
		p_0(k)\left[1 - \sum_{i=0}^{k-1} P(i)\right], & 0 < k < \text{bin}_d \\
		p_0(0), & k = 0
	\end{cases}.
	\label{eq4}
\end{equation}
Further, temporal histograms are formulated from multiple cumulative detections, which can be considered as Bernoulli trials \cite{fouche2003detection}. Therefore, the photon number detected in the $k$-th bin of the temporal histogram can be expressed as
\begin{equation}
	\tau(k) = \sum_{i=1}^{N_{\text{pulse}}} \tau_i(k), \quad \tau_i(k) \sim \text{Bernoulli}(P(k)).
	\label{eq5}
\end{equation}
where $N_{pulse}$ is the number of pulses emitted during the cumulation time. And $\tau_i(k)$ is the photon number detected in the $k$-th bin during a single pulse duration, which follows a Bernoulli distribution with a success probability of $P(k)$.

\subsection{Residual Neural Network}\label{subsec22}
Because of its strong ability to capture local correlation and maintain sequential order, 1D convolutional neural networks (CNNs) are always used as feature extractors in most traditional 1D signal processing, enhancing various tasks, such as time series analysis \cite{gamboa2017deep}, speech processing \cite{mehrish2023review}, biomedical signal processing \cite{kiranyaz20191} and motor fault diagnosis \cite{chuya2022deep}. However, due to the fluctuation caused by the Poisson process and the instability of the drone, deeper and more stable features should be extracted for drone identification. In addition, because of gradient vanishing or explosion in deep neural networks, the training process of traditional CNNs is often slow and may even fail to converge. In contrast, ResNet has shown an extraordinary ability to extract features in computer vision \cite{he2016deep} and 1D signal \cite{tchatchoua20221d} processing. In this work, instead of CNNs, we introduce the 1D version of ResNet into \text{D\textsuperscript{2}SP\textsuperscript{2}-LiDAR} for drone identification, then the meaningful features of the temporal histograms can be effectively extracted.

\begin{figure}[H]
	\centering
	\includegraphics[width=0.9\textwidth]{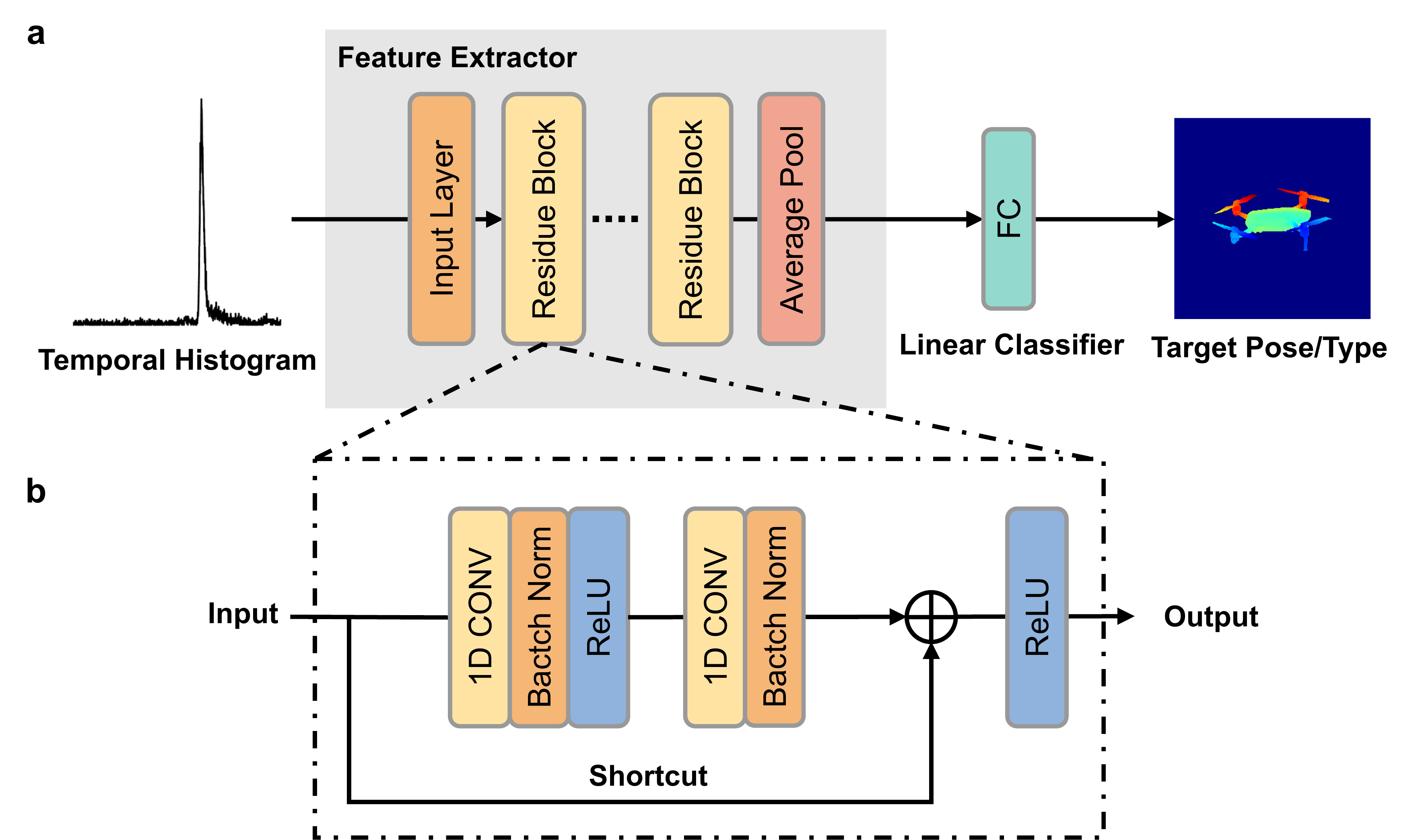}
	\caption{\textbf{a} Overall architecture of 1D-ResNet. The network is composed of a feature extractor and a linear classifier. Temporal histograms are fed to the feature extractor to generate features, then the linear classifier maps the features to the number of poses or types. \textbf{b} Detailed structure of the residual block, which is composed of two 1D convolutional blocks and a shortcut adding the input with the convolution output.}\label{fig3}
\end{figure}

The overall architecture of 1D-ResNet is shown in Fig. \ref{fig3} \textbf{a}, which consists of a feature extractor and a linear classifier. The feature extractor is constructed with a stack of residue blocks with their structure shown in Fig. \ref{fig3} \textbf{b}. The residual block is constructed from 1D convolution, batch normalization, and the Rectified Linear Unit (ReLU) activation function, with a shortcut connected from input to output. The shortcut connection helps facilitate the flow of gradients, enabling more efficient training and mitigating the issues of vanishing gradients in deeper networks. After preprocessing, which includes normalization and shift-invariant transformation \cite{hong2023image}, temporal histograms with 1024 bins are directly passed through the feature extractor to generate refined features. The features are then mapped to the probability of classes by the linear classifier, which is also a fully connected (FC) layer.

\subsection{Synthetic Data Generation}\label{subsec23}
\begin{figure}[H]
	\centering
	\includegraphics[width=0.8\textwidth]{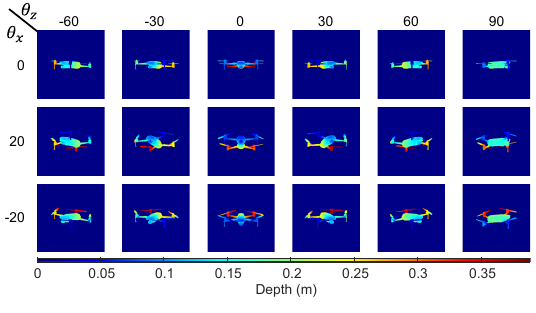}
	\caption{The generated poses of the drone model in simulation, shown as depth maps.}\label{fig4}
\end{figure}

For AI-based tasks, the dataset plays an important role and directly determines the performance of the network. But, for drone identification in long-range scenarios, the dataset is missing. Thus, to evaluate our model and train the network, we build a general model to generate the synthetic simulation dataset, in which the depth map of the target is generated by importing a 3D model into a scene, and setting up a camera within the rendering system. Then we can calculate $h(t)$ of the target based on the observation model. Considering the symmetric blur problem \cite{lai2024single} and the detection approach shown in Fig. \ref{fig2} \textbf{a}, we create a synthetic dataset with 18 different poses by rotating the target along the x-axis ($\theta_x$) and the z-axis ($\theta_z$), which is shown in Fig. \ref{fig4}. 

\begin{table}[htbp]
	\centering
	\caption{Parameters used to generate the synthetic datasets}\label{tab1}%
	\begin{tabular}{c p{10em} p{10em}}
		\midrule
		\multirow{2}[4]{*}{Parameters} & \multicolumn{2}{c}{Values} \\
		\cmidrule{2-3}    \multicolumn{1}{c}{} & \multicolumn{1}{c}{Comparison Dataset} & \multicolumn{1}{c}{Distance Dataset} \\
		\midrule
		Target distance (km) & \multicolumn{1}{c}{×} & \multicolumn{1}{c}{1, 3, 5, 6.25, 7.5, 8.75, 10} \\
		Signal noise ratio & \multicolumn{1}{c}{1, 0.1, 0.01, 0.005, 0.001} & \multicolumn{1}{c}{×} \\
		Noise level & \multicolumn{1}{c}{×} & \multicolumn{1}{c}{0.005, 0.05, 0.1, 0.5, 1} \\
		Dead time (ns) & \multicolumn{2}{c}{900} \\
		Laser pulse width (ps) & \multicolumn{2}{c}{10} \\
		Laser repetition rate (MHz) & \multicolumn{2}{c}{1} \\
		Number of pulses & \multicolumn{2}{c}{1000000, 100000, 50000, 25000} \\
		Time jitter (ps) & \multicolumn{2}{c}{150} \\
		Bin width (ps) & \multicolumn{2}{c}{10} \\
		\bottomrule
	\end{tabular}%
	\label{tab:addlabel}%
\end{table}

Then, based on $h(t)$ of each pose, we can generate temporal histograms based on Eq. \ref{eq5}. Taking practical laser pulse and SPD into account, both the time jitter $J(t)$ and the temporal shape $s(t)$ follow the Gaussian distribution. $B$ is set as a constant, which is reasonable in sunlight environments. Additionally, for a distant target, we assume that all reflective surfaces of the target share the same falloff effect coefficient, and the fluctuation of the target's position is negligible. Then, two synthetic datasets—comparison dataset and distance dataset—are generated with the parameters listed in the Tab~\ref{tab1}. 

The comparison dataset is used to evaluate the performance of different neural networks for drone pose classification, in which 100 temporal histograms with different SNR and number of optical pulses ($N_{pulse}$) are generated. Here SNR is defined as the ratio between the mean number of photons reflected from the target $N_s$ and the photon number from background noise $N_n$, that is, $SNR=N_s/N_n$. By varying $N_{pulse}$, we can generate temporal histograms with different $N_s$ and $N_n$, while varying the SNR can control the signal quality of the temporal histograms, therefore, simulating scenarios with severe background noise and weak target reflections.

The distance dataset is used to further investigate the impact of distance on neural network performance and the possibility of the system operating in long-range scenarios. We generate this dataset by positioning the drone at varying distances from 1 to 10 km. Specifically, we choose a refined interval from 5~\si{\kilo\meter} to 10~\si{\kilo\meter} to better investigate the effect of increasing distance. During generation, we assume that the drone is fully illuminated by the laser at each distance with the same illuminated area and laser power. Then the mean photon number that reaches the target can be calculated with the LiDAR equation~\cite{liu2022simulation}, which is 
\begin{equation}
	N_t=\lambda_tN_e\frac{A_t}{A_l},\label{eq6}
\end{equation}
where $\lambda_t$ represents the transmittance coefficient of the transmitter and atmosphere,  $A_t$ is the effective area of the target. For each distance, we assume that both the emitted photon numbers of the system ($N_e$) and the illuminated area of the target ($A_l$) are the same, thus $N_t$ are the same. Then, the mean photon number reflected from the target can be calculated as
\begin{equation}
	N_s=\eta\lambda_r N_t\frac{A_r}{\Omega_r d^2}=\eta\lambda_rN_t\frac{A_r}{\pi d^2},\label{eq7}
\end{equation}
where $\lambda_r$ is the coefficient containing the transmittance of the receiver, the atmosphere transmittance, and the albedo of the target. $\eta$ is the SPD detection efficiency, $A_r$ is the effective area of the receiver. $\Omega_r$ is the solid angle of the reflected light, which can be set as $\pi$ under the assumption of diffuse reflection. $d$ is the distance between the target and the SPD. Therefore, according to Eq.~\ref{eq7}, we have
\begin{equation}
	N_s=a\frac{N_e}{d^2},\label{eq8}
\end{equation}
where $a$ collectively represents the impact of the coefficients. Eq. \ref{eq8} means that we could simulate the performance of our system at a different distance by adjusting the mean signal photon count $N_s$. In simulation, we assume at the distance of 5~\si{\kilo\meter} $N_s$ is set to 5000, and $a$ is set to 1 for simplicity, then we can calculate $N_s$ in other distance conditions using Eq. \ref{eq8}. Furthermore, we introduce varying levels of noise photons to the histograms to assess the network's robustness under different noise conditions. Note that the number of noise photons is almost the same in all distance conditions, as long as the receiver effective area ($A_r$) remains the same \cite{degnan2002photon}, thus we define the noise level $N_{np}$ as the mean number of noise photons per pulse, which means that the total number of noise photons is $N_n=N_{np}N_{pulse}$. Here $N_{pulse}$ is the number of pulses emitted by the pulse laser, which is determined by the acquired time of the sample.  Similarly, for each combination of noise level and number of pulses, we generate 100 temporal histograms for training and testing. 

\section{Results}\label{sec3}
\subsection{Simulation Results}\label{subsec31}
To evaluate the performance and the ability of our \text{D\textsuperscript{2}SP\textsuperscript{2}-LiDAR} system, the comparison dataset and the distance dataset generated above are used to simulate the accuracy of drone poses classification under different conditions. The simulations here include two main tasks. One is to evaluate the performance of our 1D-ResNet network by comparing it with other neural networks under various challenging conditions, such as SNR and photon count. The other one is to test the ability of our system across a range of distances, noise levels, and number of pulses, ensuring that our system could maintain high classification accuracy even in long-range scenarios, low SNR and photon count conditions.

For the first task, the comparison of different neural networks, we conduct a ten-fold cross-validation based on the comparison dataset, in which the dataset is divided into ten subsets, one is used as a test set, while the remaining nine are used for training. This process is repeated ten times, ensuring that each subset is used as a test set exactly once. Then we average the ten results to provide a comprehensive assessment of the model's robustness and accuracy. By systematically varying SNRs and $N_s$ in the simulations, we can validate the feasibility of our approach facing real-world challenges, such as the weak echo signal returned from distant drones and potential background noise interference.

Furthermore, we construct 1D-ResNet for drone pose classification, which is implemented in PyTorch \cite{paszke2019pytorch}. The other two networks used for comparison are also implemented. One is 1D-UNet, which is implemented based on the architecture in \cite{hong2023image}, and another is the multilayer perceptron (MLP), which is constructed with three FC layers. All neural networks are trained on an NVIDIA 4090 GPU, and optimized using the adaptive moment estimation (Adam) optimizer with default parameters, learning rate 0.001, $\beta_1 = 0.9$, and $\beta_2 = 0.999$. Here, $\beta_1$ and $\beta_2$ are the exponential decay rates for the 1st and 2nd moment estimates when using the Adam optimizer. Additionally, cross-entropy is used as the loss function to guide the training process and batch size is set to 128. 

\begin{figure}[H]
	\centering
	\includegraphics[width=0.9\textwidth]{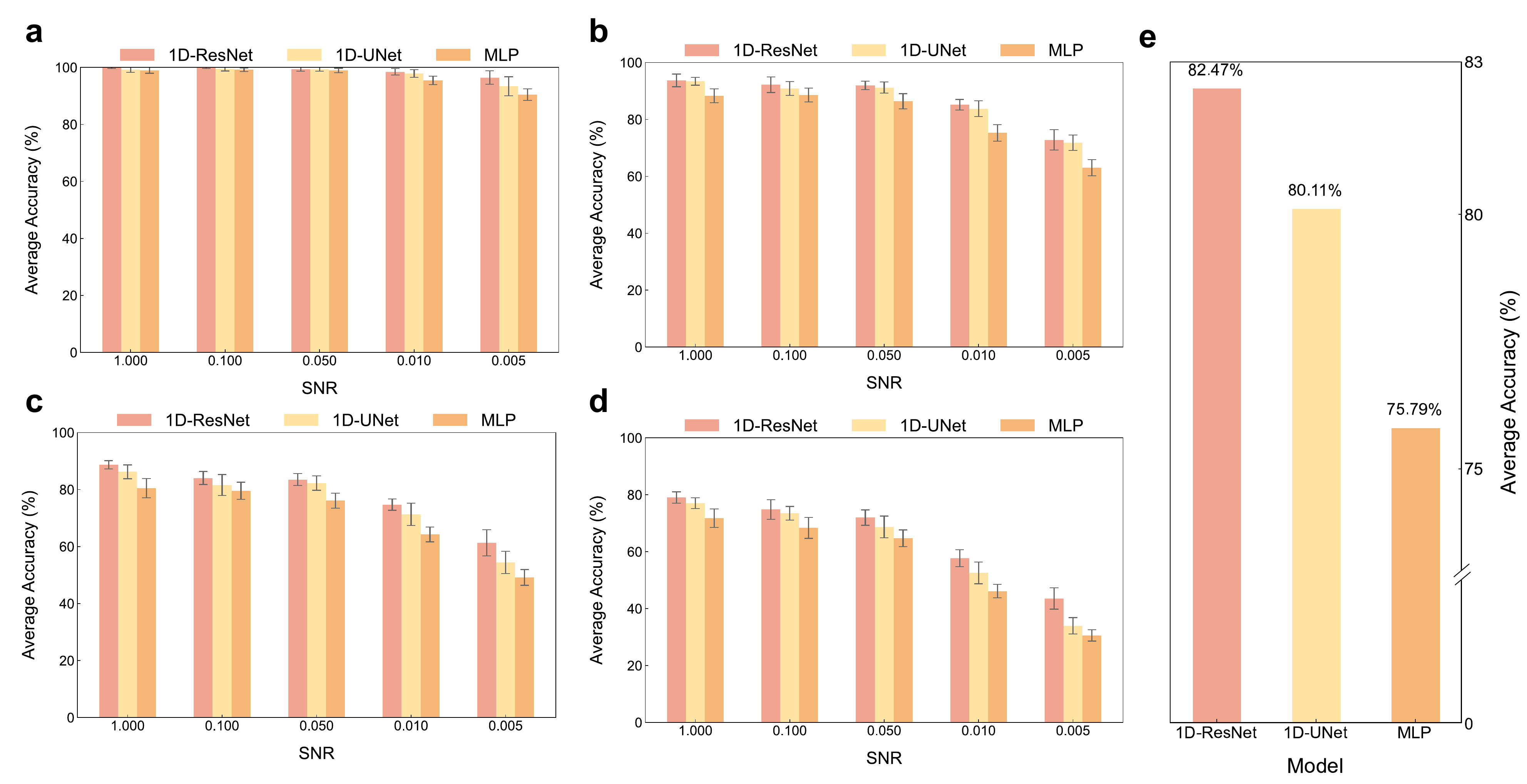}
	\caption{Average accuracy of different neural networks for 18 poses of drone model (see Fig. \ref{fig4}). \textbf{a}$\sim$\textbf{d} are results for different reflected photon number, $N_s=5000, 500, 250, 125$. \textbf{e} is the average test accuracy comparison, which is averaged under all SNRs and $N_s$.}\label{fig5}
\end{figure}

The detailed simulation results in different $N_S$ and SNRs are shown in Fig. \ref{fig5} \textbf{a}$\sim$\textbf{d}, and the average test accuracy under all SNRs and $N_s$ is listed in Fig. \ref{fig5} \textbf{e}. The results show that our 1D-ResNet outperforms MLP and 1D-UNet under all SNR and $N_s$ conditions, showing a high ability to extract more stable and useful information from temporal histograms. 
As SNR and $N_s$ decrease, the accuracy of all three neural networks decreases, which is consistent with expectations. In fact, when $N_s$ and the SNR are lower, it is difficult to distinguish the signal from the noise in the temporal histogram. However, a high average accuracy up to 82.47\% is still achieved under all SNR and $N_s$ for our 1D-ResNet network (Fig. \ref{fig5} \textbf{e}), showing the high robustness and potential of our network for practical applications in various environments. 

\begin{figure}[H]
	\centering
	\includegraphics[width=0.8\textwidth]{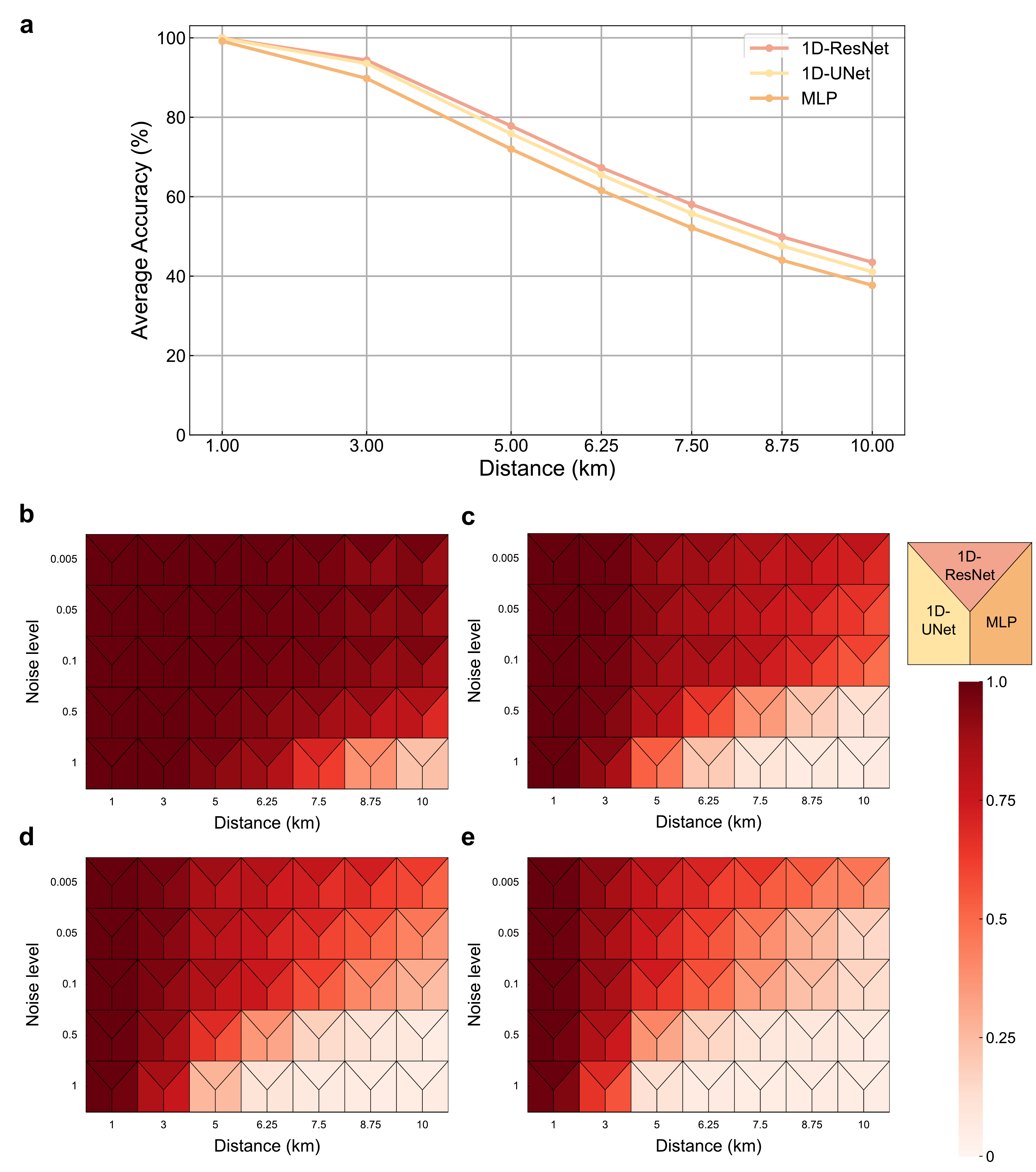}
	\caption{\textbf{a} Average accuracy of different neural networks under different distances for 18 poses of the drone model. \textbf{b}$\sim$\textbf{e} are detailed accuracy of our 1D-ResNet compared with 1D-UNet and MLP under different noise level, $N_{pulse}=1000000, 100000, 50000, 25000$ respectively.}\label{fig6}
\end{figure}

For the second task, the long-range drone detection testing, we evaluate the average accuracy of drone classification for the three neural networks mentioned above. With the distance dataset generated above, we also perform a ten-fold cross-validation to assess performance degradation across varying ranges. As shown in Fig. \ref{fig6} \textbf{a}, the average classification accuracy for 18 poses of the drone model exhibits a gradual decline with increasing distance for all neural networks, which is also expected, since the number of photons reflected from the target is significantly decreased as the distance increases. In fact, as shown by Eq. \ref{eq8}, $N_s\propto 1/d^2$, where $d$ is the distance between the system and the target. Despite this decline, our 1D-ResNet still outperforms others in all distance conditions. And it also maintains a relatively high classification accuracy at 5~\si{\kilo\meter}, and demonstrates its robustness for extended-range applications. However, beyond 5~\si{\kilo\meter}, the performance is likely to degrade further due to the significant reduction of signal photons. Fig. \ref{fig6} \textbf{b}$\sim$\textbf{e} also show the performance of our 1D-ResNet under different noise levels. As expected, when the noise level is high, the signal becomes indistinguishable from the background noise in the temporal histograms, causing a relatively low average accuracy. Thus, a longer integration time or improved signal processing techniques are required to compensate for the reduction.

\subsection{Experiment Results}\label{subsec32}
\begin{figure}[H]
	\centering
	\includegraphics[width=0.9\textwidth]{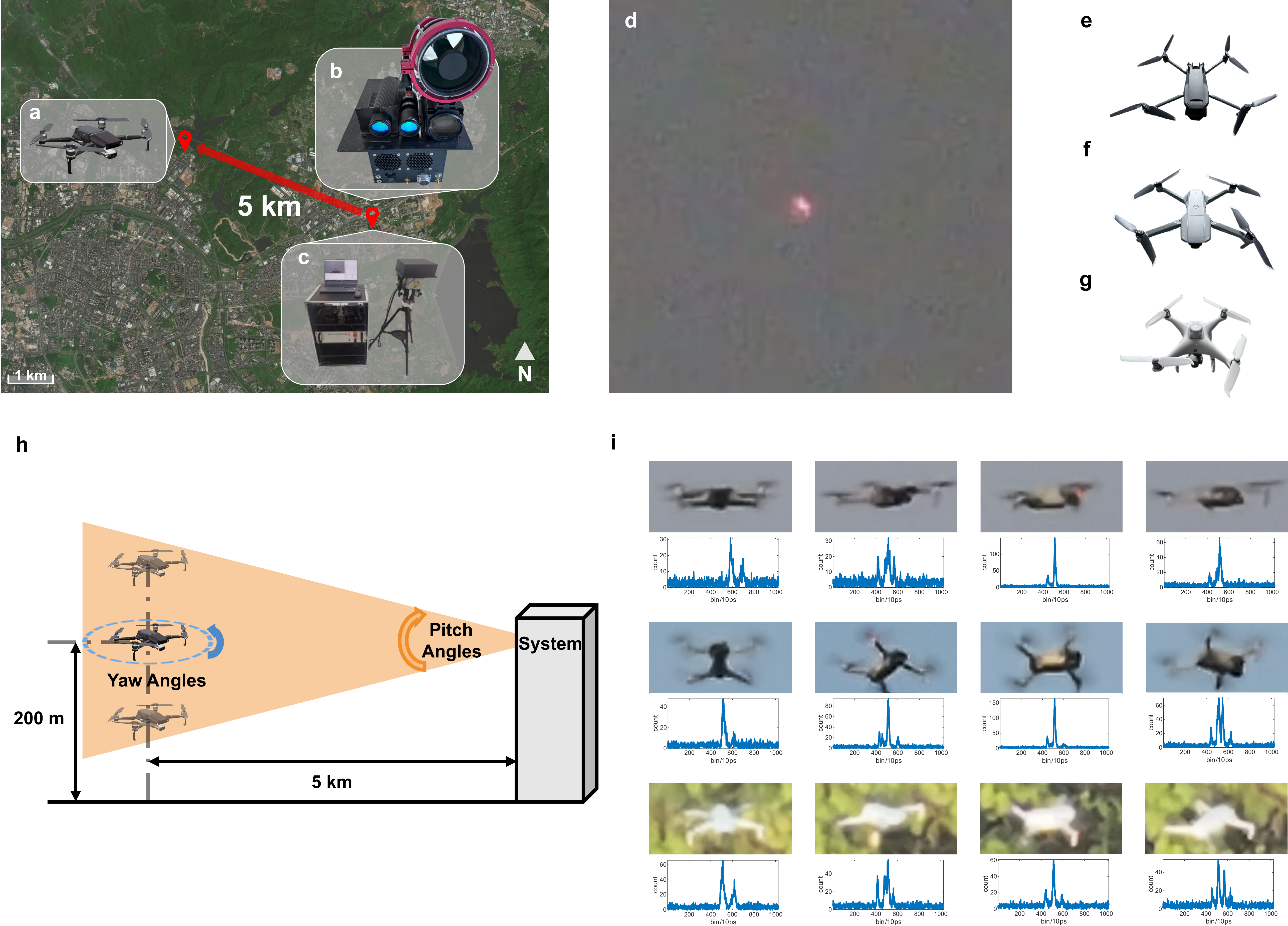}
	\caption{Real-world experiment. \textbf{a} Remote drone target at a distance of 5 kilometers. \textbf{b} \text{D\textsuperscript{2}SP\textsuperscript{2}-LiDAR} system. Two beam expanders and collimators are combined and formulated as the receiver and transmitter separately. Two cameras are used to capture and align the system to the target. \textbf{c} Overview of the system. \textbf{d} Picture of drone at the distance of 5~\si{\kilo\meter} captured by the camera of the system, whose focal length is 300~\si{\milli\meter}. \textbf{e} DJI Mavic 3 Pro. \textbf{f} DJI Mavic Air 2. \textbf{g} DJI Phantom 4 RTK. \textbf{h} Schematic of dataset collection. The drone is steered at 3 different altitudes with 4 different yaw angles and finally, formulates 12 poses. \textbf{i} Close-up photos of 12 different poses of the drone captured by another drone in daylight, and the corresponding temporal histograms acquired by the system with an integration time of 1 s.}\label{fig7}
\end{figure}

A field experiment is conducted in an intracity environment to further evaluate our method in the real world with long distance, which involves identifying 12 different drone poses and classifying three distinct types of drones. The experiment is carried out in Guangming District of Shenzhen city, from 7:00 PM to 9:00 PM, on December 17-18 and 22-24, 2024, with an average visibility of approximately 10~\si{\kilo\meter} and a temperature of roughly 16 °C. The wind speed ranges from 0.4 to 4.1 m/s and relative humidity ranges from 29\% to 48\%. As shown in Fig. \ref{fig7}, to detect a remote drone \textbf{a}, we develop a compact \text{D\textsuperscript{2}SP\textsuperscript{2}-LiDAR} system as shown in \textbf{b} and \textbf{c}. The system uses a pulsed laser with a central wavelength of 1550.13~\si{\nano\meter} and a linewidth of 0.18~\si{\nano\meter}. The average power is about 400~\si{\milli\watt} with a pulse width of 8~\si{\pico\second} and a repetition rate of 1~\si{\mega\hertz}. The transmitter and receiver are each constructed using a beam expander and a collimator, resulting in a divergence angle of 250 µrad. The reflected photons from the remote target are detected by a free-running InGaAs/InP SPAD (QCD600B from QuantumCTek) with a time jitter of 80~\si{\pico\second}, a dead time of 900~\si{\nano\second}, and a detection efficiency of 25\%. The synchronization between the laser and SPAD is achieved using TDC(Time Tagger Ultra from Swabian), whose bin width is set to 10~\si{\pico\second} in our experiment. The total time jitter of the system is approximately 100~\si{\pico\second}.

\begin{figure}[H]
	\centering
	\includegraphics[width=0.9\textwidth]{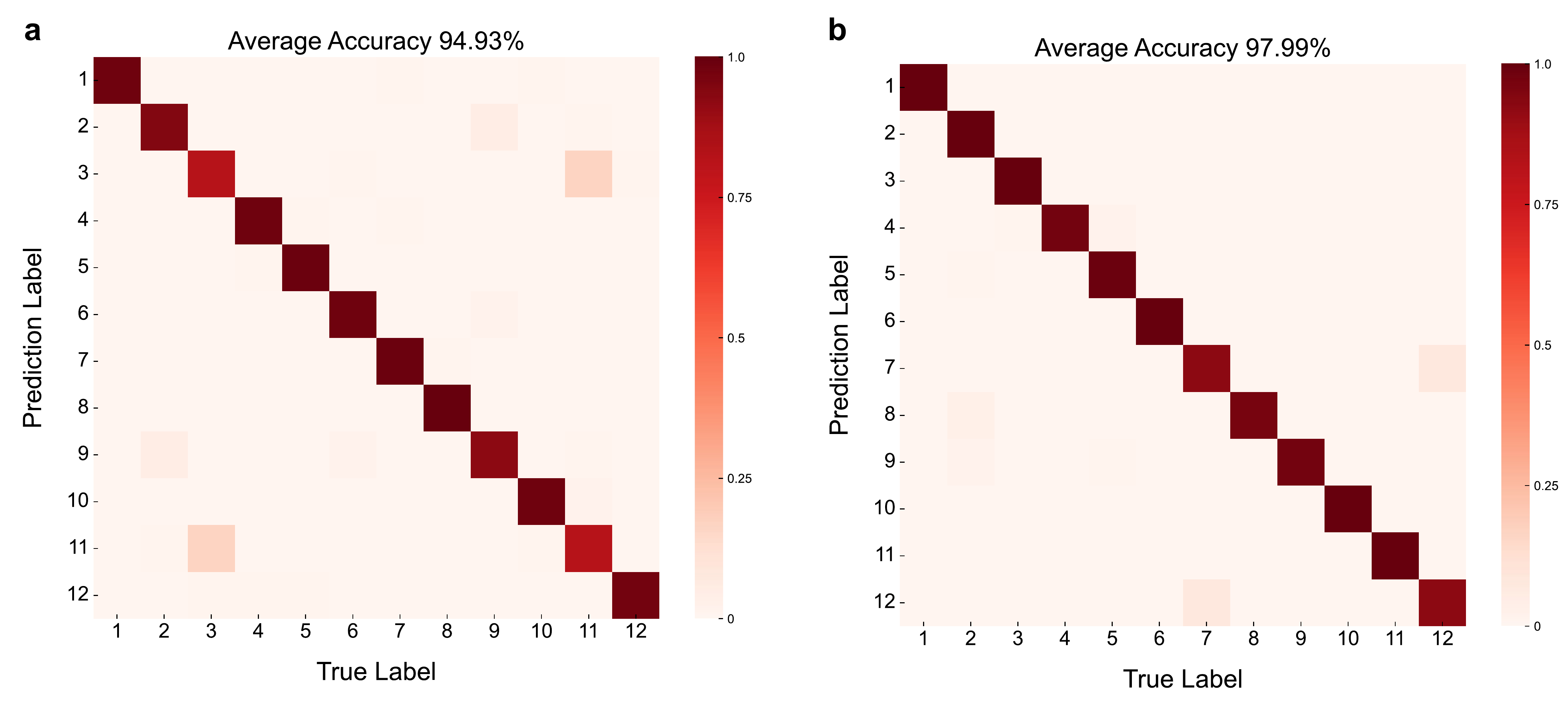}
	\caption{Real world experiment results. \textbf{a} The performance of classifying 12 different poses of DJI Mavic Air 2 at the distance of 5~\si{\kilo\meter}, shown as a confusion matrix. \textbf{b}  The performance of classifying 4 poses each for 3 different drones at the distance of 5~\si{\kilo\meter}, shown as a confusion matrix.}\label{fig8}
\end{figure}

To change the drone poses, a DJI Mavic Air 2 is controlled at three different altitudes (Fig. \ref{fig7} \textbf{h}) —100~\si{\meter}, 200~\si{\meter}, and 300~\si{\meter}—to achieve variations in pitch angles. At each altitude, four different yaw angles are set, resulting in a total of 12 distinct poses (Fig. \ref{fig7} \textbf{i}), while the corresponding temporal histograms are collected with an integration time of 1~\si{\second} (Fig. \ref{fig7} \textbf{i}). For each pose of the drone, we collect 300 temporal histograms, which are then divided into training, validation, and test sets in a 2:4:4 ratio. We train 1D-ResNet with the same hyper-parameters described in Section \ref{subsec31}, except that the batch size is changed to 32. 1D-ResNet can successfully distinguish poses with an average classification accuracy of 94.93\%, highlighting the ability of 1D-ResNet to extract meaningful features from real-world temporal histograms reflected from the target at the distance of 5~\si{\kilo\meter}. Detailed results are shown by the confusion matrix in Fig. \ref{fig8} \textbf{a}.

For the drone type classification experiment, we use a similar setup as described above, with all drones controlled at a consistent altitude of 200~\si{\meter}. For each drone type—DJI Mavic 3 Pro , DJI Mavic Air 2 and DJI Phantom 4 RTK(Fig. \ref{fig7} \textbf{e}, \textbf{f}, \textbf{g})—four different poses are set at this altitude to collect the corresponding temporal histograms. For each pose of each type of drone, we collected 300 temporal histograms, then they were split into three subsets for training, validation and test with ratio 2:4:4. The classification result is shown as a confusion matrix in Fig. \ref{fig8} \textbf{b}, showing that 1D-ResNet can precisely predict the type of drone at the distance of 5~\si{\kilo\meter} with an average accuracy of 97.99\%. For comparison, the vision picture of the drone under the same conditions can be found in Fig. \ref{fig7} \textbf{d}, which is captured by a camera with a focal length of 300~\si{\milli\meter}.

\begin{figure}[H]
	\centering
	\includegraphics[width=0.8\textwidth]{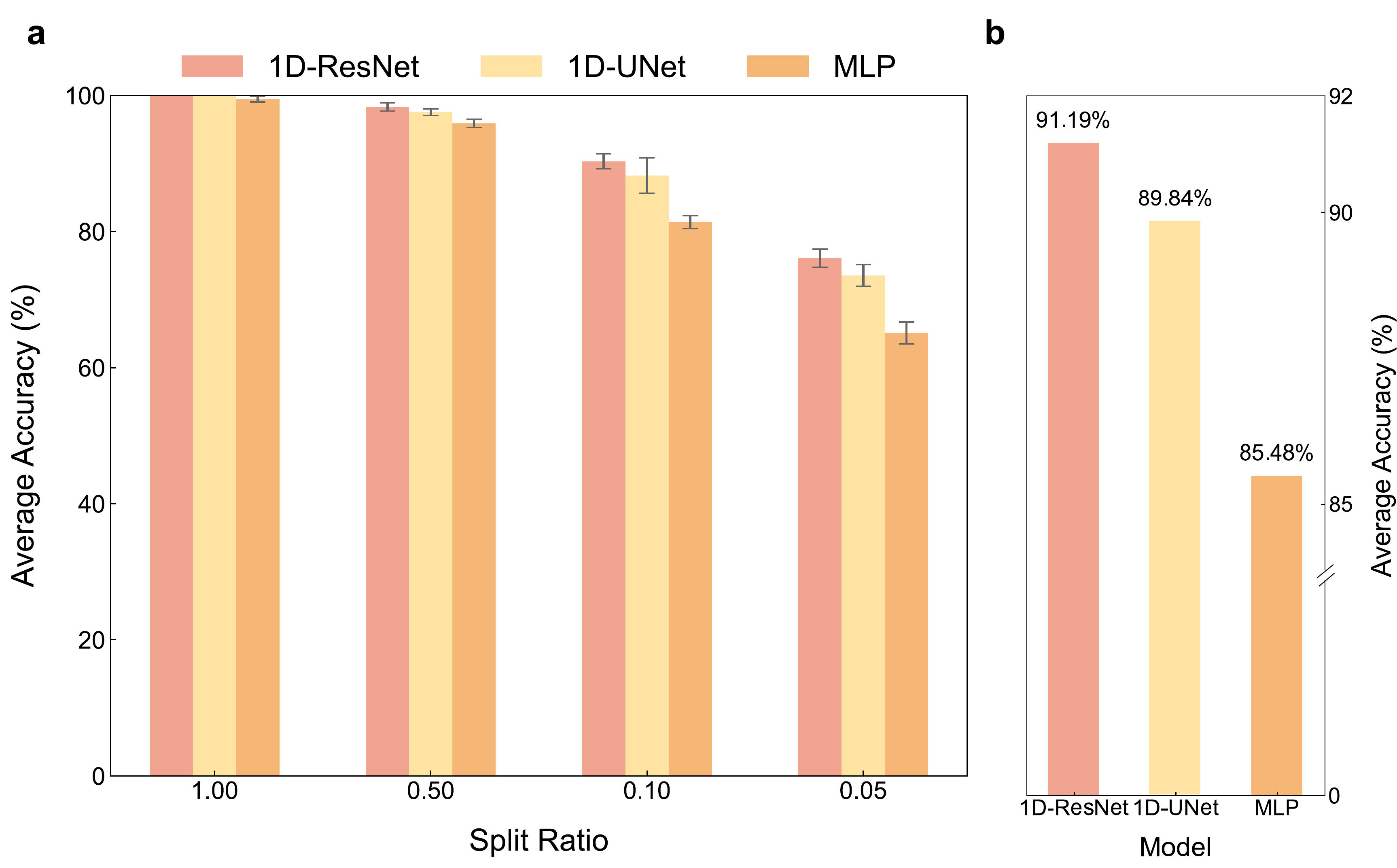}
	\caption{\textbf{a} Average accuracy of different neural networks under different split ratios. \textbf{b} The accuracy of different
		neural networks, which is averaged under all split ratios.}\label{fig9}
\end{figure}

In the experiments above, the mean photon count from the target received by SPD is around 2$\sim$5 kHz for DJI Mavic Air 2 and DJI Mavic 3 Pro, and 20 kHz for DJI Phantom 4 RTK. Thus, to further estimate the robustness of our system under low-photon count circumstances, we generate low-photon temporal histograms directly from the experimental data by performing binominal thinning \cite{hayman2020optimization}, in which binominal selection is used to split a ratio of photons from the experimental temporal histograms to create a weak version with fewer photons. Specifically, split ratios 1, 0.5, 0.1 and 0.05 are used to randomly select photons from the original histogram. Similarly, we use ten-fold cross-validation to compare 1D-ResNet with 1D-UNet and MLP. And the same hyper-parameters in Section \ref{subsec31} are used to train these networks until convergence. As shown in Fig. \ref{fig9} \textbf{a}, 1D-ResNet outperforms the other models in terms of accuracy in all split ratios, particularly under weak signal conditions, demonstrating its superior robustness, generalization, and feature extraction ability. The average accuracy of different neural networks under all split ratios is summarized in Fig. \ref{fig9} \textbf{b}. The results show an improvement in robustness and accuracy, while extending the detection range of drones to 5~\si{\kilo\meter} with \text{D\textsuperscript{2}SP\textsuperscript{2}-LiDAR}.

\section{Conclusion and Discussion}\label{sec4}
\text{D\textsuperscript{2}SP\textsuperscript{2}-LiDAR} has shown tremendous potential to detect targets in long-range scenarios while possessing the advantages of low-cost and imaging-free identification. We successfully achieve identification of drone poses and types using a \text{D\textsuperscript{2}SP\textsuperscript{2}-LiDAR} at the distance of 5~\si{\kilo\meter}, extending the detection range to new limits. Specifically, we utilize a 1D-ResNet to identify poses and types of drones based on temporal histograms collected by our compact system. The robustness and superiority in accuracy of the presented 1D-ResNet have been validated through simulation and experiment comparisons with 1D-UNet and MLP. The results indicate that in terms of robustness and accuracy, 1D-ResNet outperforms other neural networks, especially under low signal conditions. This advancement highlights the effectiveness of 1D-ResNet in extracting meaningful features from temporal histograms even in long-range scenarios, making it possible to achieve even longer detection ranges, paving the way for more accurate and robust drone detection over greater distances. However, classification under low SNR conditions requires prior information about the position of the effective signal in temporal histograms, which is intractable in real-world experiments when the signal is overwhelmed by noise. Thus, we will concentrate on developing algorithms to extract buried signal in temporal histograms. Besides, in experiments, we find that detecting drones in airspace remains challenging when relying solely on a \text{D\textsuperscript{2}SP\textsuperscript{2}-LiDAR} because of the relatively small field of view of the system. Moreover, due to the dynamic nature of drones, a relatively long integration time may introduce motion blur. Future work will focus on developing \text{D\textsuperscript{2}SP\textsuperscript{2}-LiDAR} systems that can cover a wide airspace, and algorithms to conduct faster, more accurate, and robust identification. 

Furthermore, we believe that future optimizations could potentially extend the detection range even further. For example, currently the system can reach an accuracy above 90\% with a laser pulse energy of 400~\si{\nano\J} in real-world experiments. Based on this condition, considering the distance falloff effect, we can further calculate that to detect such an amount of photons in space-borne scenarios, for example, at a distance of 500$\sim$600~\si{\kilo\meter}, will require a laser pulse energy of 4$\sim$5.76~\si{\milli\J}. In addition, implementing advanced denoising algorithms \cite{rapp2017few}, or leveraging adaptive photon thresholding \cite{chen2024adaptive} can also further push the detection range. Our findings confirm that the proposed approach is viable for long-range drone identification while also highlighting areas for potential improvement in extreme-range scenarios.

\begin{backmatter}
	\bmsection{Acknowledge} 
	The authors thank Chang Liu from QuantumCTeck for the kindly help in system building. 
	\bmsection{Funding}
	This work was supported by Shenzhen Science and Technology Program (Grant No. JCYJ20220818102014029) and National Natural Science Foundation of China (Grant No. 62171458).
	\bmsection{Disclosures}
	The authors declare no conflicts of interest.
	\bmsection{Data Availability Statement}
	Data underlying the results presented in this paper are not publicly available at this time but may be obtained from the authors upon reasonable request.
\end{backmatter}
\bibliography{sample}
\end{document}